\documentclass{article}

\usepackage{microtype}
\usepackage{graphicx}
\usepackage{subfigure}
\usepackage{booktabs} %

\usepackage{hyperref}

\usepackage[accepted]{icml2023}

\usepackage{amsmath}
\usepackage{amssymb}
\usepackage{mathtools}
\usepackage{amsthm}

\usepackage[capitalize,noabbrev]{cleveref}

\usepackage{cleveref}
\usepackage{graphicx}
\usepackage{multirow}
\usepackage{overpic} 
\usepackage{soul}
\usepackage{amsmath}
\usepackage[acronym]{glossaries}
\usepackage{booktabs}

\newcommand{\mnist}[0]{\texttt{MNIST}} %

\newcommand{\fmnist}[0]{\texttt{Fashion MNIST}} %

\newcommand{\cifart}[0]{\texttt{CIFAR-10}} %
\newcommand{\cifarh}[0]{\texttt{CIFAR-100}} %

\newcommand{\news}[0]{\texttt{N24News}} %
\newcommand{\trec}[0]{\texttt{TREC}} %

\sethlcolor{rebuttalcolor}

\newcommand{\draft}[1]{{}}  
\newcommand{\vale}[1]{{}}  
\newcommand{\ema}[1]{{}}
\newcommand{\marco}[1]{{}}
\newcommand{\antonio}[1]{{}}
\newcommand{\FL}[1]{}

\usepackage{xstring}

\newcommand{\enc}[2]{%
    \IfEqCase{#1}{%
        {abs}{absolute}%
        {rel}{relative}%
        {Abs}{Absolute}%
        {Rel}{Relative}%
        {}{}
    }[\PackageError{enc}{Undefined option to enc: #1}{}]%
    \StrLen{#1}[\encTypeLen]%
    \ifthenelse{\equal{\encTypeLen}{0}}{}{%
        \StrLen{#2}[\repLen]%
        \ifthenelse{\equal{\repLen}{0}}{}{ }%
    }%
    \IfEqCase{#2}{%
        {}{}%
        {rep}{representation}%
        {reps}{representations}%
        {Rep}{Representation}%
        {Reps}{Representations}%
    }[\PackageError{enc}{Undefined option to enc: #2}{}]%
}%

\definecolor{pltorange}{rgb}{1.        , 0.49803922, 0.05490196}
\definecolor{pltblue}{rgb}{0.12156863, 0.46666667, 0.70588235}
\definecolor{rebuttaltextcolor}{rgb}{0.6, 0.2, 0.8}
\colorlet{rebuttalcolor}{rebuttaltextcolor!25}

\theoremstyle{plain}

\theoremstyle{definition}

\theoremstyle{remark}

\usepackage[textsize=tiny]{todonotes}

\begin{document}

\twocolumn[
\icmltitle{Latent Space Translation via Inverse Relative Projection}

\icmlsetsymbol{equal}{*}

\begin{icmlauthorlist}
\icmlauthor{Valentino Maiorca}{equal,sap,ist}
\icmlauthor{Luca Moschella}{equal,sap,ist}
\icmlauthor{Marco Fumero}{sap,ist}
\icmlauthor{Francesco Locatello}{ist}
\icmlauthor{Emanuele Rodol\`a}{sap}

\end{icmlauthorlist}

\icmlaffiliation{sap}{Sapienza University of Rome}
\icmlaffiliation{ist}{Institute of Science and Technology, Austria}

\icmlcorrespondingauthor{Valentino Maiorca}{maiorca@di.uniroma1.it}
\icmlcorrespondingauthor{Luca Moschella}{moschella@di.uniroma1.it}

\vskip 0.3in
]

\printAffiliationsAndNotice{\icmlEqualContribution} %

\begin{abstract}

The emergence of similar representations between independently trained neural models has sparked significant interest in the representation learning community, leading to the development of various methods to obtain communication between latent spaces.
\emph{Latent space communication} can be achieved in two ways: i) by independently mapping the original spaces to a shared or \emph{relative}
one \citep{Moschella2022-yf}; ii) by directly estimating a transformation from a source latent space to a target one \citep{maiorca2023latent}.
In this work, we combine the two into a novel method to obtain latent space translation through the relative space. 
By formalizing the invertibility of angle-preserving relative representations and assuming the scale invariance of decoder modules in neural models, we can effectively use the relative space as an intermediary, independently projecting onto and from other semantically similar spaces.
Extensive experiments over various architectures and datasets validate our scale invariance assumption and demonstrate the high accuracy of our method in latent space translation. We also apply our method to zero-shot stitching between arbitrary pre-trained text and image encoders and their classifiers, even across modalities. 
Our method has significant potential for facilitating the reuse of models in a practical manner via compositionality.
\end{abstract}

\section{Introduction}

\begin{figure*} 
    \centering
    \includegraphics[width=.7\textwidth]{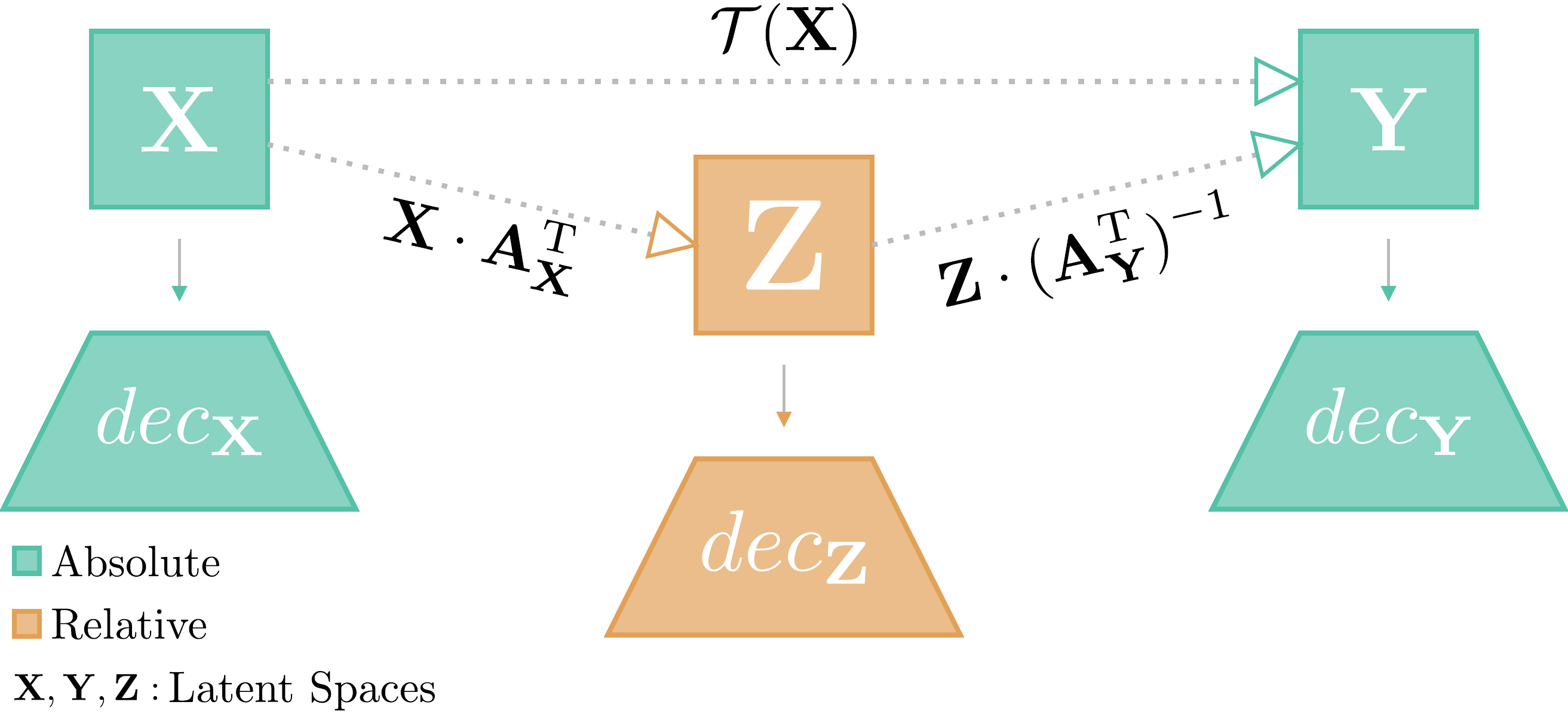}

    \caption{Zero-shot stitching of X and Y absolute spaces utilizing relative representation, direct latent translation, and our method (IRP). Relative representation requires $\text{dec}_{\mathbf{Z}}$ to stitch.
    Direct translation requires the estimation of $\mathcal{T}$ between $\mathbf{X}$ and $\mathbf{Y}$ directly, so both should be available at the same time. Instead, we first map to the bridge relative space $\mathbf{Z}$ and then, using $A_{\mathbf{Y}}$, we can independently map to $\mathbf{Y}$.}
    \label{fig:teaser}
\end{figure*}

Representation learning \citep{DBLP:journals/corr/abs-1206-5538} is a fundamental paradigm in the field of artificial intelligence, as it enables us to uncover the underlying structure of complex data. One of the main goals of representation learning is to discover a robust representation of the data that should be insensitive to certain transformations of the input, which are meaningless in order to solve the tasks of interest. The Manifold Hypothesis \citep{https://doi.org/10.48550/arxiv.1310.0425} posits that high-dimensional real-world data lies on a low-dimensional non-linear manifold: we argue that functionally equivalent models should approximate the same latent manifold \cite{Moschella2022-yf,maiorca2023latent,fumero2024latent,huh2024platonic}. However, neural models typically recover representations of the same data distribution that may differ by some transformation (e.g., a rotation), due to stochasticities in the training dynamics or extrinsic factors not related to the semantics of the data itself but should really be the same when projected on the latent manifold.
This also extends to the case of semantically similar data sampled from different distributions  (e.g., multimodal data representing the same entity). 
Identifying these transformations is critical as they hinder knowledge transfer between semantically similar latent spaces and, consequently, between neural networks. Recently, it has been observed that a transformation taking a space to a shared one, common among semantically similar spaces, can be obtained by linearly projecting on randomly selected data points (anchors) \cite{Moschella2022-yf}. Furthermore, \cite{maiorca2023latent,fumero2024latent} use the same idea of a small anchor set to directly learn a mapping between a source space and a target one.

In this paper, we propose a novel method to translate from a source space to a target one by inverting the relative representation transformation. This inversion is our core contribution since it allows us to independently map back and forth between the shared relative space. Assuming the transformation needed to perform the translation is orthogonal \citep{maiorca2023latent}, the inverse transformation method estimates the rescaling, rotation, and reflection separately, allowing us to reconstruct the absolute latent spaces from the relative representations. Through experimental evaluation of various architectures and datasets, we demonstrate the high accuracy of our method in reconstructing the absolute latent spaces. Our method also enables zero-shot stitching between arbitrary pre-trained encoders and their own classifiers without the need to concurrently access both latent spaces to fit a transformation. This makes our method a valuable tool in representation learning, allowing for the transfer of knowledge between different neural networks trained on similar data and a deeper understanding of the underlying data manifold.

\section{Related Works}

\paragraph{Representation similarity}
Recent years have witnessed a growing consensus among researchers in the deep learning community that effective neural networks tend to learn similar representations for semantically similar data, regardless of the architecture, task, or domain in which they are applied \citep{huh2024platonic}. This idea is supported by a plethora of empirical studies \citep{Moschella2022-yf,Norelli2022-ni,Morcos2018-ra,Li2015-jo,Kornblith2019-sz,Bonheme2022-tk,Tsitsulin2019-gg,Barannikov2021-eb,Vulic2020-zb,Conneau2017-vv,Lenc2014-gy,DBLP:journals/corr/MikolovLS13,NEURIPS2021_46407417,Bengio2012-ri,Movshovitz-Attias2017-rn,Chang2022-ad,cannistraci2024from,maiorca2023latent}, the phenomenon is particularly pronounced for large and wide models \citep{Somepalli2022-kw,Mehta2022-mc}. Nevertheless, despite this intrinsic similarity, latent spaces can still exhibit extrinsic variations. Our work proposes a novel zero-shot method for translating these spaces from one to another, focusing on their intrinsic similarities.

\paragraph{Stitching and zero-shot}
Model stitching, which involves the combination of different neural networks to create a new model, has been a topic of active research in the field of representation learning. A key concept in this area is that of relative representations \citep{Moschella2022-yf, Norelli2022-ni} which enables zero-shot stitching between different neural networks trained on semantically similar data.
However, this approach assumes the use of decoders trained on relative representations. To overcome this limitation and stitch arbitrary decoders, \citep{maiorca2023latent,lahner24a} directly estimate the transformation between the spaces, finding that in most cases an orthogonal one is the best suited. Our work builds upon these concepts by introducing a zero-shot mechanism for translating one absolute space to another by first projecting into the relative space and then back into the target absolute one. This enables the zero-shot stitching of arbitrarily trained models without the need for any assumptions about the decoders.
Previously, trainable stitching layers \citep{Lenc2014-gy,Bansal2021-oj,Csiszarik2021-yi} have been introduced to allow for the combination of parts of different networks or to verify statements regarding latent space similarity. 
Other works \citep{Gygli2021-qb,DBLP:journals/corr/abs-2111-07632,https://doi.org/10.48550/arxiv.2208.03861,DBLP:journals/corr/abs-2007-14906} have proposed alternative methods for obtaining compatible and reusable network components without needing stitching layers.

\paragraph{Scale invariance in neural networks}
It is well-established that neural networks exhibit positive scale-invariance in certain settings, such as when using ReLU activation functions \citep{Meng2018-te}. Additionally, several studies have analyzed the scale-invariance properties of the weights of neural networks \cite{Wang2020-ap,Li2018-gp}. Our research builds upon this understanding by providing evidence that positive scale invariance is also present in pre-trained models that are commonly used in practice. This foundational property is crucial for enabling the zero-shot translation of latent spaces, and it paves the way for efficient model reuse and combination through stitching.

\section{Method}

\subsection{Relative Representations}
Relative representation is a framework introduced in \cite{Moschella2022-yf}, enabling latent spaces of arbitrary neural models to "communicate". The method introduces an alternative way of representing samples in the latent spaces of neural networks by shifting the perspective from an absolute coordinate system to one relative to a set of predefined samples, denoted as \emph{anchors}. Specifically, the representation is computed by projecting each sample point $\mathbf{x}$ in the latent space  $\mathcal{X} \subset \mathbb{R}^k$, into the set of $\mathcal{A}_{\mathcal{X}} \subset \mathcal{X}$. Formally, this is represented as
\begin{equation} \label{eq:relreps}
\mathbf{X}_{rel} = \mathbf{X}_{abs} \cdot \mathbf{A}_{\mathcal{X}}^T\,,
\end{equation}
where $\mathbf{X}_{abs} \in \mathbb{R}^{n \times d},\mathbf{A}_{\mathcal{X}} \in \mathbb{R}^{k \times d}$ and $\mathbf{X}_{rel} \in \mathbb{R}^{n \times k}$, with $n=|\mathcal{X}|,k=|\mathcal{A}_\mathcal{X}|$. Samples in $\mathcal{X}$ and in $\mathcal{A}$ are rescaled to unit norm, i.e. $\mathbf{x}=\frac{\mathbf{x}}{\| \mathbf{x} \|_2} \ \ \ \forall x \in \mathcal{X}$ and $\mathbf{a}=\frac{\mathbf{a}}{\|\mathbf{a}\|_2} \ \  \ \forall \mathbf{a} \in \mathcal{A}_{\mathcal{X}}$. This projection captures the intrinsic shape of the data by only considering the \textbf{relative angles} between points, thus completely losing the scale information.

One of the main contributions of relative encoding is that it shows how the signal encoded in the angles is enough to represent the information, reaching results on various benchmarks comparable to those using absolute encodings. Furthermore, they demonstrate that different latent spaces, in practice, differ only by an isometry transformation plus local rescaling if they share the same data semantics. 
This is also shown in \cite{maiorca2023latent} where Procrustes analysis is efficiently used to map with high performance from space $\mathbf{X}$ to $\mathbf{Y}$,
The discovery that a rigid transformation is enough to map between the two is crucial for our work as it allows us to estimate the translation, rescaling, rotation, and reflection constituting this isometry separately, \textit{enabling the decoding of a single relative space into different absolute spaces}. While the original relative representations method assumes that "NNs commonly employ normalization techniques...to center the latent spaces around zero", we add a centering step to enforce it since we do not rely on any training that could mitigate the impact of not centered spaces.

\subsection{Latent Space Translation}
The key benefit of the relative projection is its non-injective nature, as it maps different absolute spaces into a single relative space. The core of our method lies in the formalization of the inverse process by exploiting the information contained in the anchor points. At a high level, this means we can use the relative space as a middle ground to translate an encoding from an absolute space $\mathcal{X}$ to any other semantically similar absolute space $\mathcal{Y}$. This can be formalized as:
\begin{equation} \label{eq:rel2abs}
\mathbf{Y}_{abs} = \mathbf{X}_{rel} \cdot (\mathbf{A}_{\mathcal{Y}}^T)^{-1}\,,
\end{equation}
under the assumption that $\mathbf{X}_{rel} \approx \mathbf{Y}_{rel}$.
This transformation of relative to absolute coordinates allows us to transfer any encoding between absolute spaces using only a subset of points from both spaces (the anchors). The anchors of the source space, the \textit{encoding space} $\mathcal{X}$, are used to project into the shared relative space, while the anchors of the target one, the \textit{decoding space} $\mathcal{Y}$, are used to project back to the specified absolute one.

As with the other "latent space communication" methods, the anchor points must be in semantic correspondence to effectively translate between the spaces. In fact, they represent the only bridge, a partial correspondence between them, much like a ``Rosetta stone'' \citep{Norelli2022-ni}.

\subsection{Stability improvements}
In this section, we present techniques for enhancing the stability of the process for inverting the anchors and then evaluate their impact.
Consider the anchor matrix $\mathbf{A} \in \mathbb{R}^{n\times d}$, where $n$ is the number of anchors. A condition for computing its (pseudo) inverse is the linear independence of its columns. In a synthetic setting, where the vectors in $\mathbf{A}$ can span the whole ambient space $\mathbb{R}^d$, the curse of dimensionality \citep{curse} makes it highly unlikely that two vectors in $\mathbf{A}$ will be linearly dependent. However, in practical scenarios, $\mathbf{A}$ is composed of only a reduced set of parallel anchors, and it is difficult to arbitrarily expand it using the entire training set \citep{cannistraci2023boot, vedula2024scalable}. 
Additionally, the manifold hypothesis restricts the subspace where the encodings live, which is not the whole ambient space. This makes it more likely for two random points to be similar enough to negatively impact the inverse process's feasibility.
This is the reasoning behind the following proposed stability improvements. 

\paragraph{Anchor Pruning}
Given a set of anchors, denoted as $\mathcal{A}$, which are randomly drawn from the training data and have a number of samples $k$ equal to $d$ itself, our goal is to refine this set to enhance the stability of the inverse process. As a subset of the data manifold, the anchors may exhibit high correlation, leading to many singular values during the inverse computation. To mitigate this, we introduce a technique called ``anchor pruning''.

We employ the greedy farthest point sampling (FPS) algorithm~\citep{fps} to select a subset $\mathcal{S}$ of $\mathcal{A}$ with maximum orthogonality. The distance metric we use for FPS is a customized cosine distance, calculated as:
\begin{equation} \label{eq:dcos}
     \mathrm{dcos} (\mathbf{X}) =   1 - \bigg|\dfrac {\mathbf{X} \cdot \mathbf{X}^T} {\left\| \mathbf{X}\right\|^2 }\bigg|\,,
\end{equation}
where $| \cdot |$ denotes the element-wise absolute value to avoid quasi-colinear points with opposite directions. A stopping condition ($\delta$) is applied based on the minimum acceptable distance between points. The subset of anchors selected by FPS is defined as:
\begin{equation} \label{eq:fps}
    \mathbf{S} = \mathrm{FPS}(\mathbf{A}, \mathrm{dcos}, \delta)\,,
\end{equation}
where $\delta$ is the minimum acceptable distance between points. The remaining vectors in $\mathbf{S}$ span a {subspace} of the original anchor set $\mathbf{A}$.
This process has the effect of selecting a highly informative set of anchors while reducing the correlation between them, which improves the stability of the inverse process.

\paragraph{Anchor Subspaces}
To balance the reduction in the cardinality of the anchor set given by the anchor pruning, we apply the pruning procedure multiple times with a different random seed by controlling it with the parameter $\omega$. For each subspace $S^i$, we independently reconstruct the absolute space. Then, we average pool the reconstructed spaces in a parameter-free ensemble fashion:
\begin{equation} \label{eq:subpool}
    \mathbf{Y}_{abs} =
    \frac{1}{\omega}\times
    \sum_{i=1}^{\omega}{\mathbf{X}_{rel} \cdot (\mathbf{S}_{\mathcal{Y}}^{i\,T})^{-1}} \,.
\end{equation}
This approach allows us to maximize the coverage of the original set of anchors while maintaining stability in the inverse process. It can be seen as an ensembling step, where the absolute encodings are reconstructed as the mean of the predictions of different estimators. 

Overall, this process increases the method's robustness to the stochastic factors introduced by the anchor pruning while also mitigating information loss.

\paragraph{Anchor Completion}
The objective of this step is to get back to a set of $d$ points as anchors, such that the dimensionality of the points is not altered when transforming $\mathcal{X}_{abs}$ into $\mathcal{X}_{rel}$. To achieve this, we first encode the identity matrix using any anchor subset $\mathcal{S}^i_x$ and decode it using the  anchor set $\mathcal{A}_y$. This process reveals how the canonical basis of $\mathbf{X}$ maps to relative space and back to space Y. The canonical basis in $\mathbf{X}$ can now be considered the new set of anchors for X (denoted as $\mathcal{A}^{}_{X}$), and the corresponding points as the new set of anchors for Y (denoted as $\mathcal{A}^{}_{\mathcal{Y}}$). Intuitively, this process: i) stabilizes the computation without altering the information encoded by the anchors, just better distributing it across different dimensions; ii) is semantically equivalent to the estimation of the rotation and reflection matrix that transforms $\mathbf{X}$ into $\mathbf{Y}$ since the relative encoding of $\mathbf{X}$ with respect to the new $\mathbf{A}^{}_{\mathcal{X}}$ is $\mathbf{X}$ itself.

With this, we obtain back a number of anchor points equal to $d$ so that when we project X into the relative space, we don't change the dimensionality of the points.

\subsection{Scale invariance and temperature}\label{sec:scale-invariance-temperature}

 \begin{figure} 
    \centering
    \begin{overpic}[trim=1.25cm 2cm 1 3cm,clip,width=0.85\linewidth]{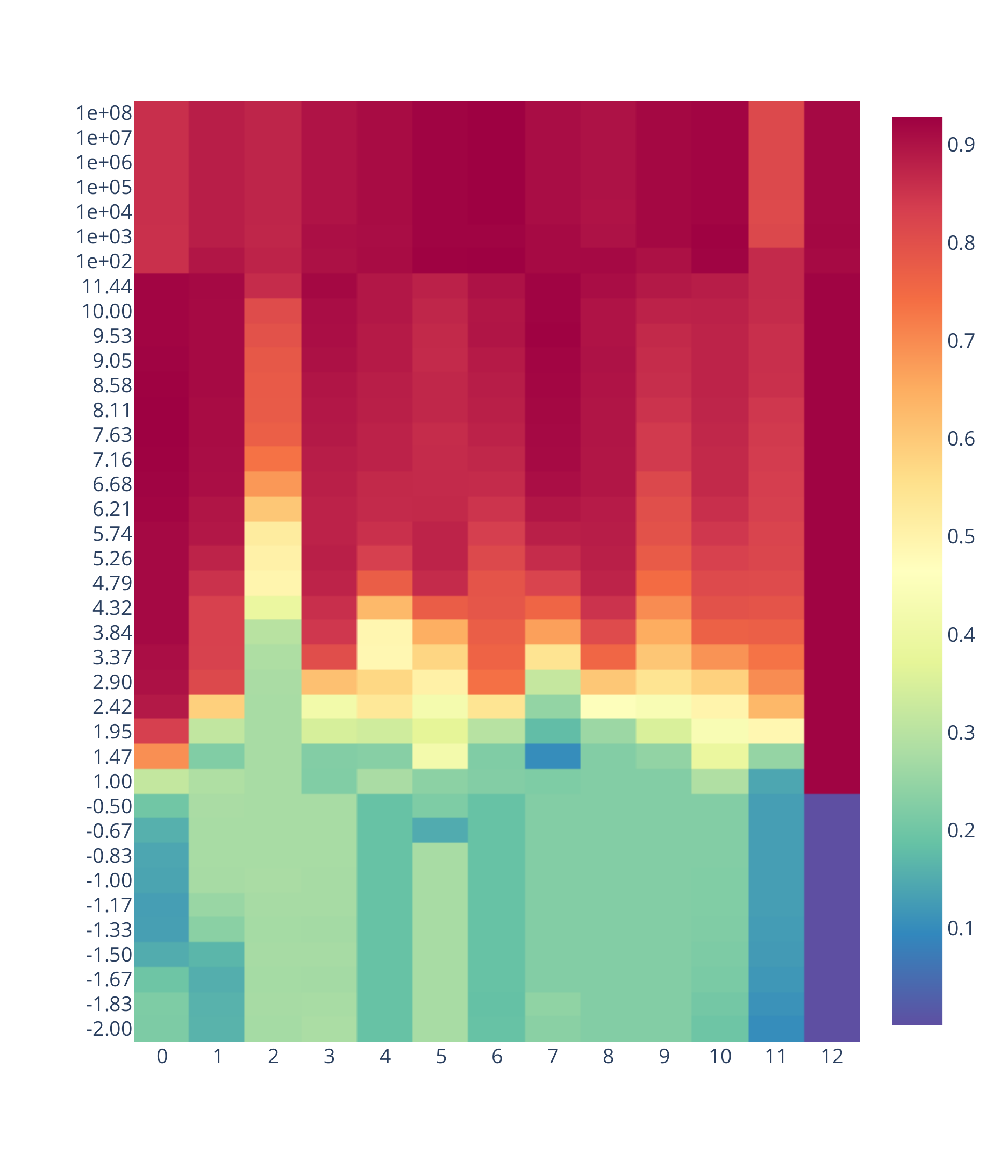}
        \put(33, 0){Rescaled layer}
        \put(0, 35){\rotatebox{90}{Rescaling factor $\alpha$}}
        \put(93, 59){\rotatebox{-90}{Accuracy}}
    \end{overpic}
    \caption{Scale invariance of RoBERTa according to the performance of a downstream classifier trained on the encodings of the last attention layer. At each layer (with 0 being the embedding layer and 12 the output one), one for each run, we rescale the encodings by the specified $\alpha$ and measure its effect on the final accuracy. The performance without any rescaling is $0.92$. }
    \label{fig:rescaled-layer-acc}
\end{figure}

Here, we aim to formally investigate the scale-invariance properties of neural classifiers that utilize the softmax activation function. We focus on the effect of rescaling operations on the latent input encodings and demonstrate that, by construction, certain classifiers exhibit scale-invariance properties without the need for additional priors.   

The softmax function, commonly used in neural classifiers, is known to be a temperature-controlled variant of the maximum function:
\begin{equation}
\operatorname{softmax}(x)_i=\frac{e^{\frac{y_i}{T}}}{\sum_j^N e^{\frac{y_j}{T}}}\,.
\end{equation}

This means that the softmax temperature can be used to control the level of confidence of the classifier's predictions. In this study, we show that a similar effect can also be achieved by rescaling the latent encodings given as input to a trained (and frozen) classifier.

In order to show this, we first note that the rescaling factor, $\alpha$, can be factored out of the matrix multiplication in the Linear layers of the classifier. This can be represented mathematically as:
$\mathbf{y} = \alpha \mathbf{W}\mathbf{x} + b$, where $\mathbf{x}$ is the input latent encoding, $\mathbf{W}$ is the weight matrix, $b$ is the bias vector, $\alpha$ is the rescaling factor, and $\mathbf{y}$ is the output of the linear layer. This implies that the rescaling operation can be ``pushed through'' the classifier without affecting its final prediction as it becomes equivalent to some temperature value applied at the softmax level.

Furthermore, we investigate the effect of rescaling when non-linear activation functions are involved and posit that as long as the function has a monotonic interval, if we rescale all the dimensions by an amount similar to the mean scale of the encodings on which the classifier was trained, we end up in the monotonic interval, without losing the scale-invariance property. In \Cref{sec:scale-distribution}, we link the monotonic interval of activation functions to the scale range of the data encodings.

In summary, our study provides formal evidence that neural classifiers that utilize the softmax activation function can, in practice, maintain their scale-invariance properties when the input latent encodings are rescaled. This property is essential to our method, as it allows us to ignore the exact scale when decoding back to the absolute space.

\section{Relative Inversion}\label{sec:relative-inversion}
In this section, we evaluate the capabilities and effectiveness of our proposed method. We first demonstrate the results of inverting a relative space by decoding it back into the original absolute space on which it was computed. We'll refer to this as \textit{intra-space inversion}. This is a valuable illustration of the method's effectiveness and components. Building upon this, we leverage the property that the relative representations of semantically similar spaces are relatively consistent, generalizing to \textit{inter-space inversion}. This enables decoding into an absolute space that is different from the one used for the relative encoding, effectively enabling latent translation.

\paragraph{Experimental setting}
All the performed experiments contain a relative inversion pattern: 
given a dataset $D$ and a set of encoders ${E_i}$, where $i = 1, 2, ..., n$, our experiments aim to translate a sample $x$ from the latent space produced by $E_i$ to the latent space produced by $E_j$. The translation consists of the following steps: (1) centering and L2 normalization; (2) relative encoding with the anchors embedded by $E_i$; (3) relative decoding with the anchors embedded by $E_j$; (4) de-normalization and de-centering according to the anchor statistics. The dimensionality of $E_i$ and $E_j$ is arbitrary and in general, it may differ, i.e. we can zero-shot translate latent space of different dimensionalities.

\subsection{Space Inversion}
We evaluate the performance of our proposed method by analyzing its sensitivity to its two hyperparameters: $\omega$ (number of subspaces) and $\delta$ (pruning threshold). Specifically, we aim to understand how the choice of these parameters affects the overall performance of the method and identify the optimal values for each. 

\paragraph{Vision}
We consider \cifarh{}, an image classification dataset, and embed each contained image with 5 different encoders: 3 vision transformers \citep{vit}, RexNet \citep{rexnet}, and the visual encoder of CLIP \citep{clip}. For each sample and for each encoder, we then apply the relative encoding followed by the relative inversion with different values for $\omega$ and $\delta$. We are interested in: i) measuring the \textit{reconstruction similarity} in terms of cosine similarity (since we can ignore the scale) between the original encoding and the reconstructed one; ii) whether $\omega$ and $\delta$ affect the performance in the same way for both the intra-space and inter-space settings. We use 768 anchors and group by all the results by $\omega$ and $\delta$.

\begin{figure} 
    \centering
    \begin{overpic}[trim=0 -1.5 0 -2.5ex,clip,width=0.9\linewidth]{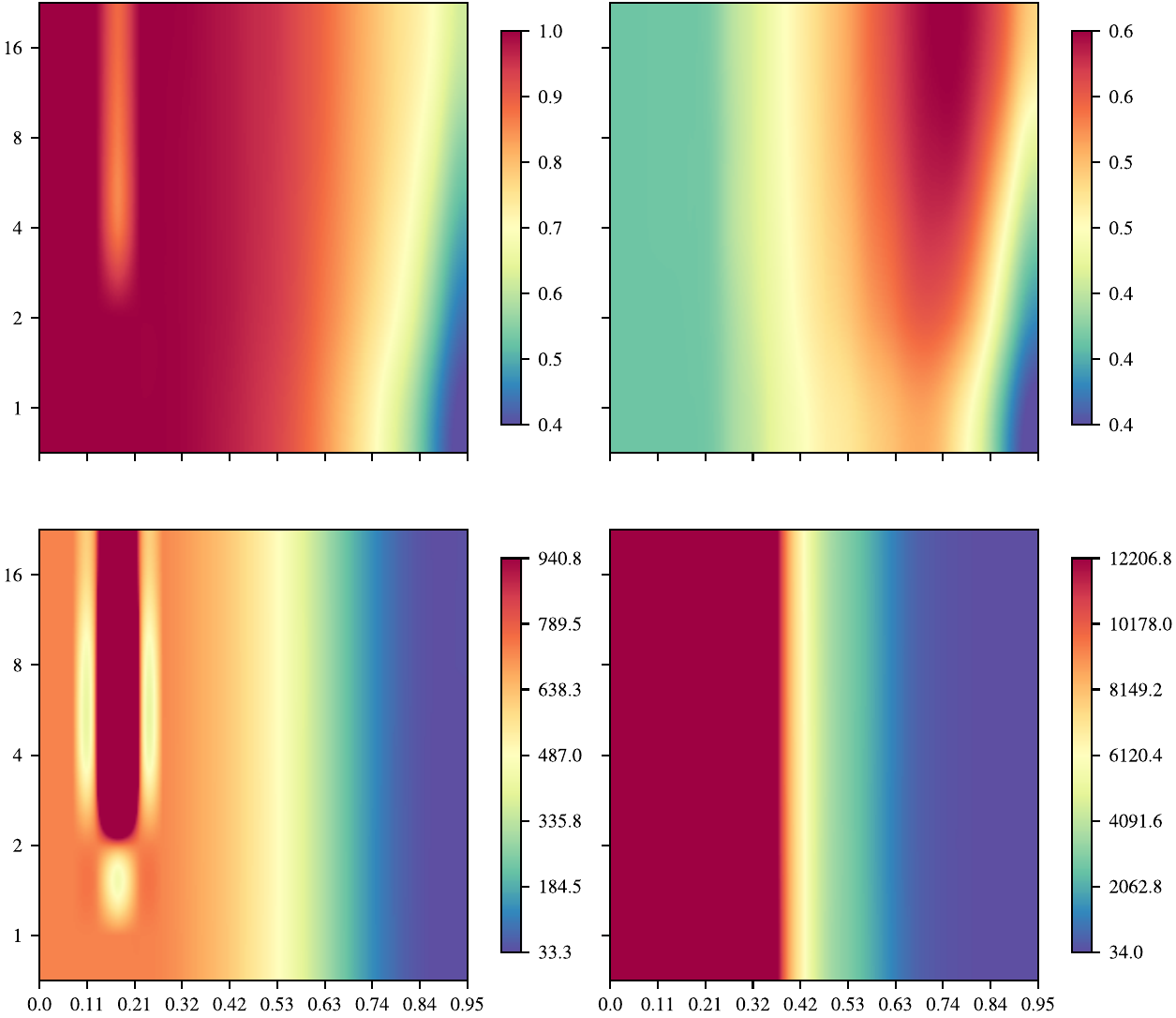}
        \put(62, 89){\small Inter space} 
        \put(13, 89){\small Intra space}

        \put(-3, 58){\rotatebox{90}{\scriptsize \# Subspaces ($\omega$)}}
        \put(-3, 12){\rotatebox{90}{\scriptsize \# Subspaces ($\omega$)}}

        \put(100.5, 84){\rotatebox{-90}{\scriptsize Reconstruction Similarity}}
        \put(100.5, 35){\rotatebox{-90}{\scriptsize Condition Number}}

        \put(8, -3){\scriptsize Pruning Threshold ($\delta$)}
        \put(57, -3){\scriptsize Pruning Threshold ($\delta$)}
    \end{overpic} 
    \caption{On the top row, reconstruction similarity sensitivity at different number of subspaces and pruning threshold ($\delta$) of intra-space inversion (\textit{left}) and inter-space inversion between different encoders (\textit{right}) on the coarse-grained Cifar100. On the bottom row, the corresponding condition number average over the subspaces. Higher pruning thresholds lower the condition number, stabilizing the matrix inverse, thus increasing the reconstruction similarity.}
    \label{fig:inner-space-inversion}
\end{figure}
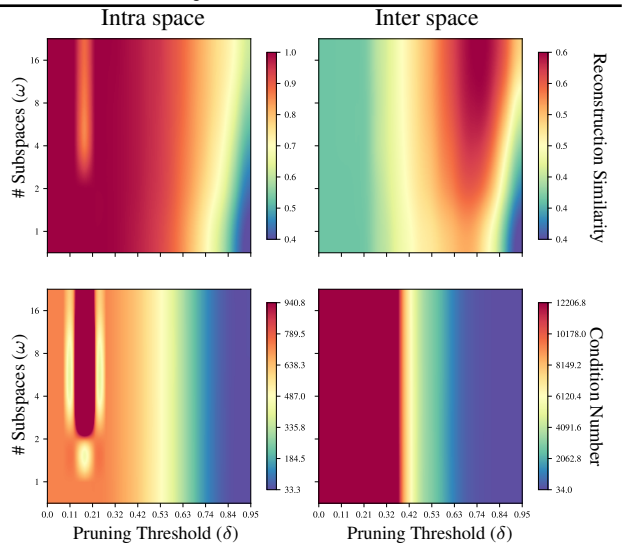

The results in \Cref{fig:inner-space-inversion} show that: i) the inverse is working since we obtain medium to high similarity between the original encoding and the reconstructed one; ii) $\omega$ has little to no impact in the intra space setting when the number of anchors pruned is low (low $\delta$), but is really helpful in compensating the effects of the aggressive pruning (high $\delta$); iii) the inter-space setting highlights the synergy between the two, where the balance is found in a medium-high pruning but with high resampling (high $\omega$).

\paragraph{Language}
We consider WikiMatrix \citep{wikimatrix} as a source for parallel multi-lingual data. It consists of translations of the same sentence in different languages automatically extracted from Wikipedia. We select ~4000 samples with a high probability of being different translations of the same sentence in 4 languages. For each language, we select a language-specific RoBERTa-based encoder: English \citep{roberta}, Spanish \citep{es_roberta}, French, and Japanese. We encode each sentence in a sample using its corresponding language-specific encoder and a language-agnostic one, specifically XLM-R \citep{xlmr}. We frame our task as translating a mono-lingual encoding in language $x$ to its corresponding mono-lingual encoding in language $y$. We select the cosine similarity between the two as a score metric and apply it to compare: \textit{i)} the decoded representation of our method, Inverse Relative Projection (IRP); \textit{ii)} the multi-lingual representations of XLM-R; \textit{iii)} the original mono-lingual encodings of the two languages. 

The results in \Cref{tab:cross-lang-inversion} show that our latent translation method works across languages. 
Interestingly, the mean \texttt{en} similarity between all possible different pairs of different embeddings in the absolute and XLM-R space are respectively $0.93 \pm 0.05$ and $0.98 \pm 0.03$. 

These experiments demonstrate that the reconstruction is effective across various domains (including different languages) and with different encoders. Additionally, it highlights that using a smaller number of anchor sets is often the optimal choice for capturing the nuances of the target space.

\begin{table} 
\centering
\caption{Reconstruction similarities between parallel encodings across languages. \textit{Absolute} refers to the original encodings, \textit{XLM-R} to the encodings of a multilingual LLM, \textit{IRP} to our Inverse Relative Projection, where we compare the translated source embeddings with the target ones. 
}
\label{tab:cross-lang-inversion}
\begin{tabular}{llccc}
\toprule
\texttt{Enc.}  & \texttt{Dec.}  &  \texttt{Absolute}  &  \texttt{XLM-R}  &  \texttt{IRP}  \\
\midrule
\texttt{en} & \texttt{en} &             1.00 &              1.00 &            0.97 \\
\texttt{es} & \texttt{en} &             0.00 &              0.99 &            0.97 \\
\texttt{fr} & \texttt{en} &            -0.02 &              0.99 &            0.97 \\
\texttt{ja} & \texttt{en} &             0.03 &              0.99 &            0.97 \\
\bottomrule
\end{tabular}
\end{table}

\subsection{Scale invariance}

In this section, we delve into the concept of scale invariance in neural networks and its implications for model compositionality. By examining the behavior of networks when subjected to a specific type of input manipulation,  \textit{rescaling injection}, we aim to demonstrate the robustness and versatility of neural networks in handling different scales of input data, a property that enhances the applicability of our IRP.

\paragraph{Rescale Injection}

\begin{figure} 
    \centering
    \begin{overpic}[trim=0 0 0 0, clip,width=.9\linewidth]{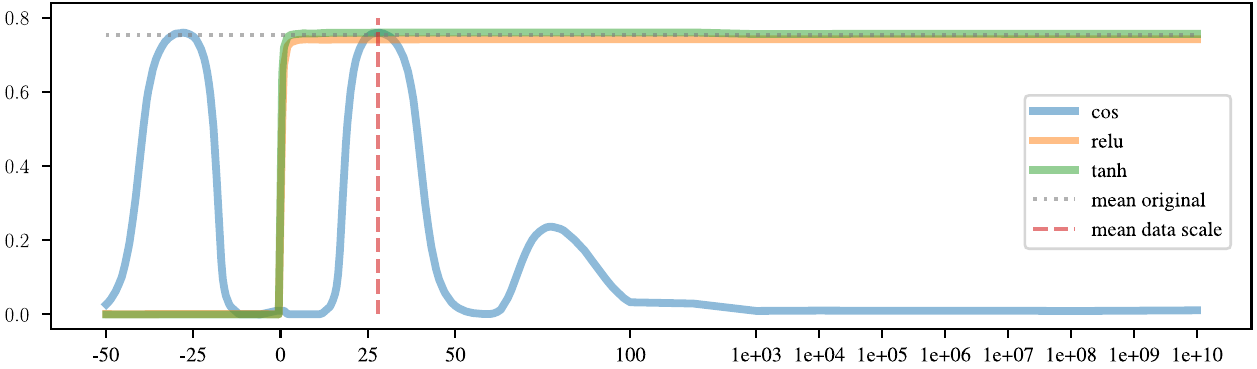}
        \put(-3.5, 7){\rotatebox{90}{\scriptsize Accuracy}}
        \put(40, -3){\scriptsize Rescale Factor}
    \end{overpic}
  
    \caption{Performance comparison of three Multilayer Perceptrons (MLPs) with different activation functions, namely cosine (blue), ReLU (orange), and tanh (green) at different rescaling factors $\alpha$. The ReLU and tanh MLPs exhibit scale invariance, while the cosine activation function is only invariant on the mean data scale and its periodic cycles. 
    }
    \label{fig:scale-inv-mono}
\end{figure}

We define the \textit{rescaling injection} as the operation of artificially altering the scale of the features produced at a specific layer of the network. We achieve this by normalizing the embeddings to the unit norm and then rescaling them by a factor of $\alpha$. By varying the value of $\alpha$, we can observe how the network's performance is affected at different scales. Through this empirical analysis, we aim to provide insight into the scale invariance properties of neural networks and their potential for use in model compositionality.

In \Cref{fig:scale-inv-mono}, we present experimental results regarding this investigation. We trained simple multi-layer perceptrons (MLPs) composed of two hidden layers, with no normalization layers, using encodings produced by the Clip Vision transformer (\texttt{clip-vit-base-patch32}) on the \cifarh{} (fine) dataset. The MLPs were evaluated using different activation functions: cosine (blue), tanh (orange), and ReLU (green). The rescaling injection technique was applied directly to the input embeddings, setting their norm to $\alpha$. 

We can observe that the scale of the embeddings does not significantly impact the MLPs' performance when using monotone activation functions that do not flip signs. This is a non-trivial result, as the nonlinearity of the activation function, the presence of bias terms $b$, and the absence of normalization layers make it difficult to predict the performance impact of rescale injection. 
It is particularly interesting to see that the cosine activation function shows an oscillatory performance comparable to the original embeddings when rescaled by the mean embeddings scale (vertical red line) or its opposite since it is symmetric.

Our findings indicate that, surprisingly, even the internal layers of large deep learning models exhibit a \textit{positive scale invariance}, as illustrated in Figure \ref{fig:rescaled-layer-acc}. The underlying mechanism for this behavior is not straightforward, but we hypothesize that it may result from the interplay between various factors, such as the choice of activation function, the use of normalization layers, the optimization objective and regularization techniques employed during the training phase. Even though further research is needed to fully understand and explain this phenomenon, we can exploit it for our inverse projection purposes since it allows us to ignore the encoding scale.

\begin{figure} 
    \centering
    \begin{overpic}[clip,width=0.95\linewidth]{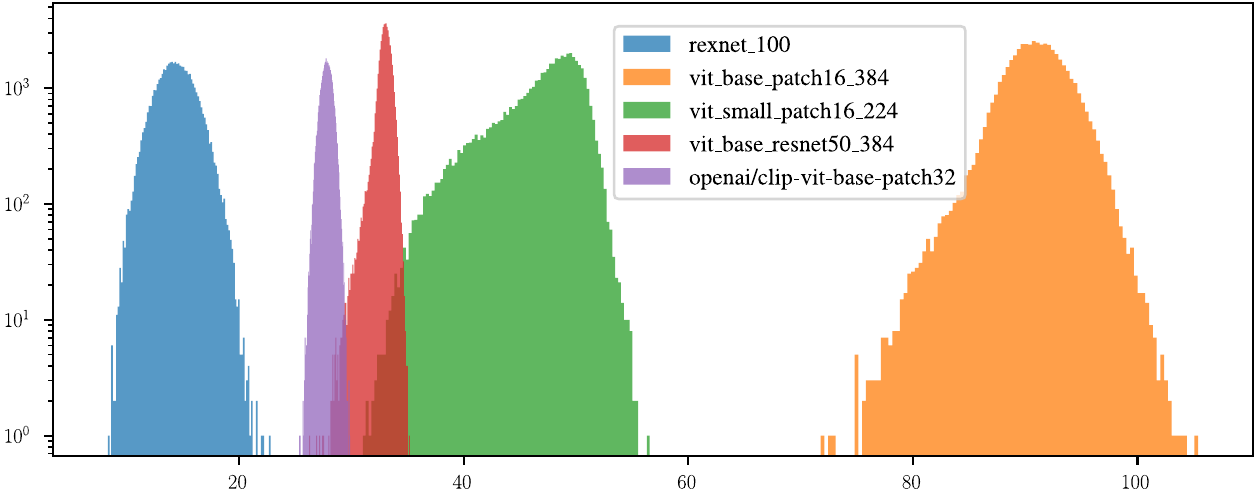}
        \put(-3, 4){\rotatebox{90}{\scriptsize Number of samples ($\log$)}}
        \put(45, -3){\scriptsize Scale}
    \end{overpic}
    \caption{Distribution of the embedding scales in different pre-trained encoders on the Cifar100 dataset. Well-behaved Gaussian distributions with a single mode and well-defined mean are crucial for our method, as they support the ability to rescale the embeddings to a mean scale.}
    \label{fig:norm-ranges}
\end{figure}

\paragraph{Scale distributions}\label{sec:scale-distribution}
In \Cref{fig:norm-ranges}, we present the scale distribution of the embeddings produced by several encoders on the \cifarh{} (fine) dataset. 
This empirical analysis shows a consistent pattern among encoders, in that the scale distribution of their embeddings follows a Gaussian one with a single mode and a well-defined mean, with minimal presence of outliers. This consistent behavior across encoders is likely attributed to their architectural choices, such as the normalization techniques, regularizations and the optimization problems they are designed to solve. 

Overall, the scale invariance property and the encoders' well-behaved scale distribution enable the zero-shot translation between arbitrary latent spaces via our IRP method.

\section{Relative space as a bridge}
\begin{table*} 
\centering
\caption{\textbf{Zero-shot stitching for classification}. Comparison between stitching and non-stitching accuracies. Stitching includes our translation method (\textit{zero-shot}) with its hyperparameters and the stitching of \textit{absolute} spaces. The \textit{achievable} rows show the number of possible combinations the stitching methods can tackle among the \textit{available} ones. Our method works between encoders of arbitrary embedding dimensionality thus it can always stitch together all the available pairs. The results are averaged over 5 different random seeds. The (\textit{non-stitch}) column reports the performance of models trained end-to-end for reference.}
\label{tab:stitching}
\begin{tabular}{llcccccc}
\toprule
  & & &    &  \multicolumn{3}{c}{Accuracy}  & \\
   \cmidrule(lr){5-7}
  & & $\omega$ & $\delta$ & zero-shot &  absolute & non-stitch & \textit{available} \\
\midrule
\multirow{6}{*}{\rotatebox{90}{Vision}} & \mnist{} & 8 & 0.65 & $0.88 \pm 0.03$ & $0.10 \pm 0.03$ & $0.97 \pm 0.01$ & 20 \\
& \fmnist{} & 8 & 0.65 & $0.82 \pm 0.03$ & $0.10 \pm 0.02$ & $0.91 \pm 0.01$ & 20 \\
& \cifart{} & 8 & 0.75 & $0.92 \pm 0.05$ & $0.11 \pm 0.04$ & $0.95 \pm 0.03$ & 20 \\
& \cifarh{} \textit{(coarse)} & 8 & 0.70 & $0.72 \pm 0.10$ & $0.05 \pm 0.01$ & $0.88 \pm 0.05$ & 20 \\
& \cifarh{} \textit{(fine)} & 8 & 0.75 & $0.62 \pm 0.09$ & $0.01 \pm 0.00$ & $0.82 \pm 0.07$ & 20 
\\[1.15ex]
& \textit{achievable} &  & & 20 & 6 &  & \\
\midrule
\multirow{3}{*}{\rotatebox{90}{NLP}} & \trec & 8 & 0.70 & $0.73 \pm 0.09$ & $0.22 \pm 0.06$ & $0.87 \pm 0.04$ & 20 \\

& \news{} & 8 & 0.65 & $0.55 \pm 0.10$ & $0.05 \pm 0.01$ & $0.70 \pm 0.07$ & 20 
\\[1.15ex]
& \textit{achievable} &  & & 20 & 20 &  & \\
\bottomrule
\end{tabular}
\end{table*}

In this section, we combine the inter-space translation capabilities (\Cref{sec:relative-inversion}) of our method and the scale-invariance of neural classifiers (\Cref{sec:scale-invariance-temperature,sec:relative-inversion}) to obtain zero-shot stitching of independently pre-trained models and classification heads. It should be noted that this method is distinct from previous approaches that rely on relative representations, as our method does {\em not} assume that the decoders have been independently trained on relative representations. 
Additionally, since the relative space itself is the bridge between absolute ones, the transformation is still independent and not conditioned on the target space as in \cite{maiorca2023latent}. In fact, each absolute space gets \textit{independently encoded} to the relative one, and each absolute space can be \textit{independently decoded} just by selecting appropriate anchors. This makes our approach more versatile and applicable to a wider range of scenarios, especially when the target space is unknown beforehand.

\paragraph{Experimental setting} We consider a variety of Computer Vision (\mnist{} \citep{mnist}, \fmnist{} \citep{fmnist}, \cifart{}, \cifarh{}  \cite{Krizhevsky2009-hv}) and Natural Language Processing (\trec{} \citep{trec} and \news{} \citep{wang-EtAl:2022:LREC3}) datasets. For the text domain, we consider 4 different language models as encoders, and for the image domain, 5 encoders, all pre-trained and frozen.
For each dataset and encoder, we train a classification head on top of their specific encodings. We then measure the mean performance over all the combinations of (encoder, classification head) for each test set in different settings: i) \textit{zero-shot}, this is the result of the application of our inter-space translation, for which we also report $\omega$ and $\delta$; ii) \textit{absolute}, this is the result of using the encodings without any transformation, we consider this as a probe for any pre-existing compatibility of encodings; iii) \textit{non-stitch}, the performance of the classification head applied to the original space it was trained on. We consider this as an upper bound. 

Lastly, we select one dataset, \trec{}, to zoom into the sensibility to our hyperparameters $\omega$ and $\delta$, considering their impact over: \textit{i)} the reconstruction similarity; \textit{ii)} the accuracy on the downstream task; \textit{iii)} the mean condition number across the subspaces; \textit{iv)} the mean number of anchors after pruning across subspaces. 

\paragraph{Result analysis}
The complete stitching results are in \Cref{tab:stitching}. As expected, the \textit{absolute} encodings achieve a score comparable to random guessing while also considering fewer encoder combinations out of the possible total due to the dimensionality discrepancy between some of them. 

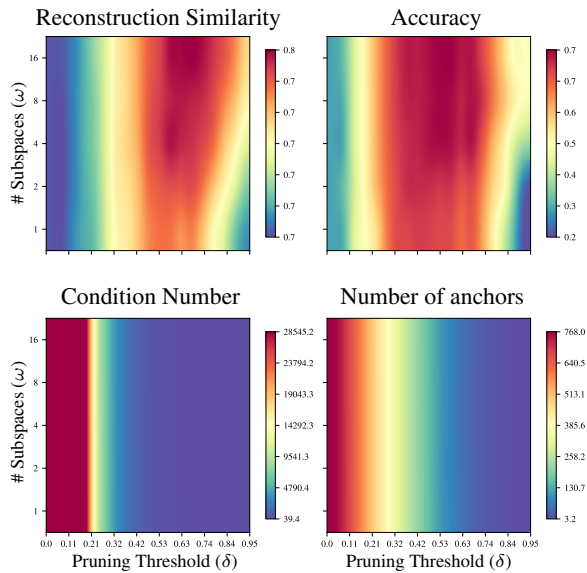
\begin{figure}[h] 
    \centering
    \begin{overpic}[trim=0 0 0 -4ex, width=0.9\linewidth]{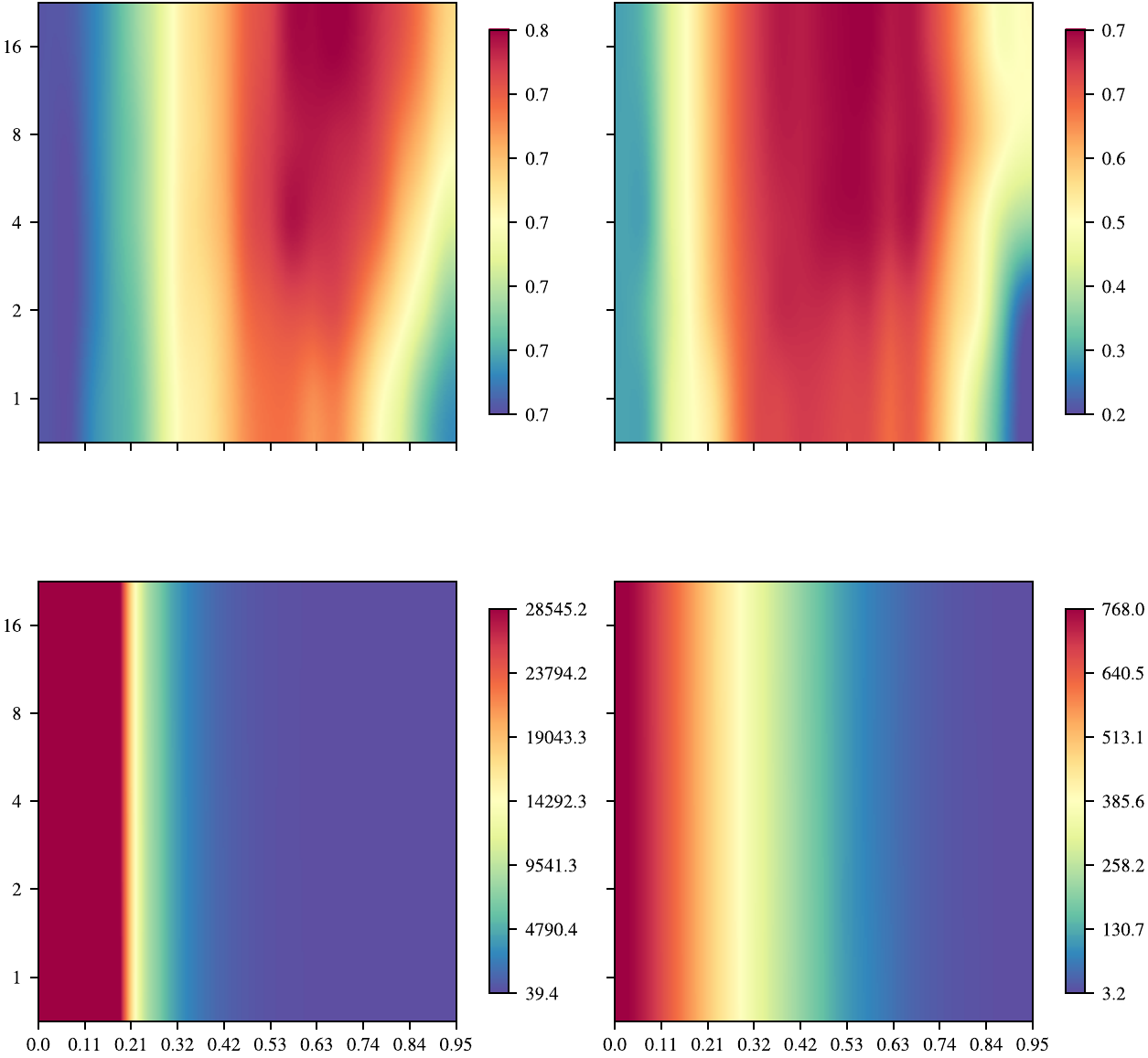}
        \put(1.5, 94){\small Reconstruction Similarity}
        \put(65, 94){\small  Accuracy} 

        \put(6, 44){\small Condition Number}
        \put(56, 44){\small  Number of anchors}

        \put(-3, 62){\rotatebox{90}{\scriptsize \# Subspaces ($\omega$)}}
        \put(-3, 12){\rotatebox{90}{\scriptsize \# Subspaces ($\omega$)}}

        \put(8, -3){\scriptsize Pruning Threshold ($\delta$)}
        \put(57, -3){\scriptsize Pruning Threshold ($\delta$)}
    \end{overpic} 
    \caption{On the top row, the reconstruction similarity (\textit{left}) and accuracy (\textit{right}) sensitivity analysis on the number of subspaces ($\omega$) and pruning threshold ($\delta$) on the \trec{} dataset. On the bottom row, the corresponding condition number (\textit{left}) and the number of anchors after pruning (\textit{right}) averaged across subspaces.  
    We report the average metrics from stitching between all possible pairs of language encoders across 3 seeds. }
    \label{fig:stitching-trec}
\end{figure}

Additionally, as expected, \Cref{fig:stitching-trec} shows a high correlation between the reconstruction similarity and the classification accuracy: their Pearson correlation is $0.7$. 
Interestingly, performance-wise, having a stable inverse (i.e., a low condition number) is more important than many anchors, which intuitively should better represent the data. Indeed, the best performance is obtained with few anchors and a low condition number. Moreover, our anchor pruning technique successfully stabilizes the inverse, as demonstrated by the high Pearson correlation between the condition number and the number of anchors ($0.93$).

This result shows the effectiveness and stability of our inversion method in zero-shot classification scenarios, as well as its ability to maximize the information extracted from the starting anchor set.

\subsection{Cross-domain stitching}
As a last experiment, we test the capabilities of our inversion method to translate representations between distinct domains in a zero-shot fashion.

\begin{figure*} 
    \label{fig:cross-domain-stitching}
    \centering
    \begin{minipage}{.7\columnwidth} 
        \begin{overpic}[trim=-1cm -1cm 0 0,width=1\linewidth]{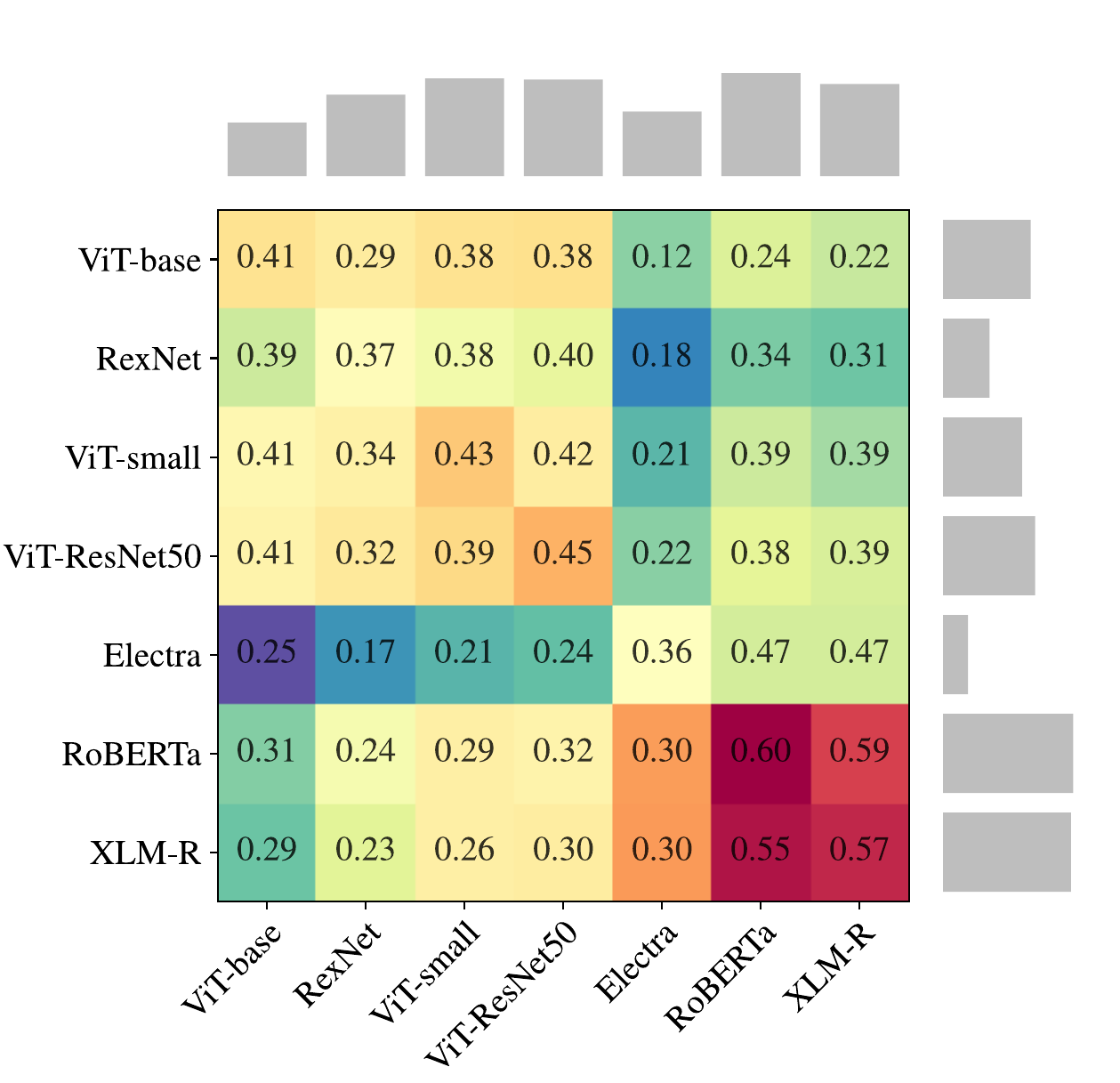}
            \put(-0.25, 41){\rotatebox{90}{\scriptsize Decoder}}
            \put(43, 0){\scriptsize Encoder}
        \end{overpic} 
    \end{minipage}
    \begin{minipage}{.29\columnwidth}
     \scriptsize
         \setlength{\tabcolsep}{0pt}
        \begin{tabular}{p{3ex}p{12ex}p{6ex}}
        \toprule
              & Model & Acc \\
        \midrule
        \multirow{4}{*}{\rotatebox{90}{Vision}}    &  ViT-base &  0.50 \\
            &   RexNet &  0.42 \\
            & ViT-small &  0.48 \\
            & ViT-ResNet &  0.50 \\[1.5ex]
        \multirow{3}{*}{\rotatebox{90}{NLP}}    &    Electra &  0.60   \\
            &   RoBERTa &  0.76 \\
            & XLM-R &  0.71 \\
        \bottomrule
        \end{tabular}

    \end{minipage}

    \caption{Performance comparison between different encoders and data modalities on  the \news{} multimodal dataset. On the right the performance of models trained end-to-end on a single data modality. On the left the stitching performance between pairs of encoders and decoder. The stitching achieves satisfactory results even when translating across modalities. Results obtained with $1000$ anchors, $\omega=16$ and $\delta=0.8$.}
\end{figure*}

\paragraph{Experimental setting}
We select \news{} as in \cite{maiorca2023latent}, since it is a multimodal news classification dataset that contains both text and associated pictures. We use different pre-trained uni-modal encoders to apply the standard encoding procedure to these two features separately. Then, we train a classification head on top of each one. Lastly, we stitch each encoder with a classification head different from its corresponding one, measuring its classification accuracy. 

\paragraph{Result analysis}
The results are in \Cref{fig:cross-domain-stitching}. The discrepancy in the mean accuracy represented by the marginal bar plots is a signal that can be used to identify spaces more suited to be \textit{decoded into} and the ones that are stronger in \textit{encoding from}. In fact, the language models as source space for the translation exhibit stronger performance than the vision encoders, as observed by \cite{maiorca2023latent}. We relate this behavior to the higher generality of the text domain data used during pre-training with respect to the image domain one \citep{Zhai_2022_CVPR}.

Overall, these results show that our method can effectively zero-shot translate across different domains.

\section{Conclusion}
In this work, we have proposed a new method for performing \textit{zero-shot latent space translation}.
At the heart of our proposed method is the synergy between three key elements: the formalization of an inverse transformation from the relative space to an absolute one, the scale-invariant properties of deep neural networks, and the Gaussian distribution of the embedding scales. Together, these elements form the foundation of our approach and enable the zero-shot translation between different spaces in the latent space of deep neural networks. This inverse transformation opens up many applications, and we chose to focus on the stitching of arbitrarily trained models on semantically similar data for classification purposes. To this end, we show how it can be used to zero-shot re-use neural network modules in different pipelines, even when they have been trained independently.

Our results not only address the zero-shot stitching problem for classification but also showcase the potential of this method for a wide range of applications in the deep learning field. We believe this work has the potential to become a step forward in the ability to re-use pre-trained models and is a strong contribution to the research area of model compositionality.

\paragraph{Future works and limitations}
As with any new approach, there are limitations that warrant further exploration of our proposed method. For example, the optimal number of anchor points required for different tasks and datasets to boost performances, investigating the factors that could be linked to latent space compatibility (e.g., their intrinsic dimension), trade-offs between the granularity of the anchor set and its condition number, alternative methods for computing relative representations when parallel anchors are not available.

These are exciting research directions that we believe hold great potential for advancing the field and improving the effectiveness and robustness of our proposed method.

\bibliography{references}
\bibliographystyle{icml2023}

\end{document}